\documentclass[10pt,twocolumn,letterpaper]{article}

\usepackage{iccv}
\usepackage{times}
\usepackage{epsfig}
\usepackage{graphicx}
\usepackage{amsmath}
\usepackage{amssymb}
\usepackage{subfig}

\usepackage{booktabs}

\usepackage[pagebackref=true,breaklinks=true,letterpaper=true,colorlinks,bookmarks=false]{hyperref}

\iccvfinalcopy 

\def\iccvPaperID{26} 

\ificcvfinal\fi

\usepackage[accsupp]{axessibility}
\begin{document}

\title{The Marine Debris Dataset for Forward-Looking Sonar Semantic Segmentation}

\author{Deepak Singh\\
Netaji Subhas Institute Of Technology\\
Dwarka Sec-3, Delhi, India\\
{\tt\small deepaks.ec.16@nsit.net.in}
\and
Matias Valdenegro-Toro\\
German Research Center for Artificial Intelligence\\
Robert-Hooke-Str 1, 28359 Bremen, Germany\\
{\tt\small matias.valdenegro@dfki.de}
}

\maketitle
\ificcvfinal\thispagestyle{empty}\fi

\begin{abstract}
   Accurate detection and segmentation of marine debris is important for keeping the water bodies clean. This paper presents a novel dataset for marine debris segmentation collected using a Forward Looking Sonar (FLS). The dataset consists of 1868 FLS images captured using ARIS Explorer 3000 sensor. The objects used to produce this dataset contain typical house-hold marine debris and distractor marine objects (tires, hooks, valves,etc), divided in 11 classes plus a background class. Performance of state of the art semantic segmentation architectures with a variety of encoders have been analyzed on this dataset and presented as baseline results. Since the images are grayscale, no pre-trained weights have been used. Comparisons are made using Intersection over Union (IoU). The best performing model is Unet with ResNet34 backbone at 0.7481 mIoU.
\end{abstract}

\section{Introduction}
The advancements in robotics and artificial intelligence have led to robots being used for a lot of applications in daily lives. One of those applications is cleaning. Robots like iRobot's Roomba are being a lot for household cleaning and Excavators for coarse waste recycling \cite{icinco21}, but a large portion of the seafloor is also polluted with marine debris that needs cleanup \cite{valdenegro2016submerged}. The extensive progress in deep learning and computer vision has enabled better scene understanding, object detection and segmentation capabilities for robots on land and air. A major contributing factor to this is the availability of large datasets for training the deep learning models for segmentation of objects. State of the art semantic segmentation architectures use encoders whose weights have been trained on large datasets like ImageNet, CityScapes, etc.

For underwater robots, object segmentation and detection still remain a challenge. Due to heavy scattering and attenuation, optical information gathered by a camera cannot be used for precise segmentation of underwater objects. This raises the need for other sensors to understand underwater environments and the most common choice is a sound navigation and ranging system (SONAR).

A forward looking sonar (FLS) is a better option than an optical camera in most scenes as it is independent of water turbidity and optical visibility. FLS can capture highly detailed images of underwater scenes at a high frame rate (approximately 15 Hz). But issues such as acoustic shadows and reflections make the interpretation of these images difficult. One reason for slow progress is lack of a proper FLS dataset for training deep neural network models. Although there have been previous works like \cite{uw_place_recog} which proposed an underwater place recognition technique using FLS sensor to detect loop closures, there has not been any significant progress in public datasets for segmentation of sonar images. Aykin et al. \cite{aykin2012feature} and Haghighat et al. \cite{haghighat2016segmentation}  used K-means for segmentation of FLS images, while Reed et al \cite{reed2004automated} segments images using a Markov Random Field, and some image classification techniques \cite{myers2010template} require shadow and highlight segmentation information. While  Marine debris in underwater environments is a good target for detection and segmentation \cite{iniguez2016marine}, as it is target for removal or extraction, and has a large inter and intra class variability. There are datasets for segmentation of some marine debris classes \cite{hong2020trashcan}, but only in color images which is not optimal.

In order to tackle the lack of datasets for training deep learning models for semantic segmentation on FLS images, this paper presents a novel dataset collected using a forward looking sonar (FLS) for marine debris segmentation. The data is collected using ARIS Explorer 3000 sensor. There are 1868 images in the dataset with 11 object classes and a background class. Pixel-wise annotations have been made on the images to create the ground truth labels for training deep convolutional network models. State of the art semantic segmentation architectures like Unet \cite{ronneberger2015u}, PSPnet \cite{zhao2017pyramid}, LinkNet \cite{chaurasia2017linknet} and DeepLabV3 \cite{chen2017rethinking} have been tested on this dataset. Since most of the available encoders are pre-trained on RGB images and the images in this dataset are grayscale, the models have been trained from scratch on this dataset. These models give the baseline results for semantic segmentation on this dataset. Mean Intersection over Union (mIOU) parameter has been used to compare the results.

The detailed explanation for data collection, labeling and training the baseline models are given in further sections.

\section{Marine Debris Segmentation Dataset}

\subsection{Data Collection}
\begin{figure}[!t]
    \centering
    \includegraphics[scale=1.0, width=0.35\textwidth]{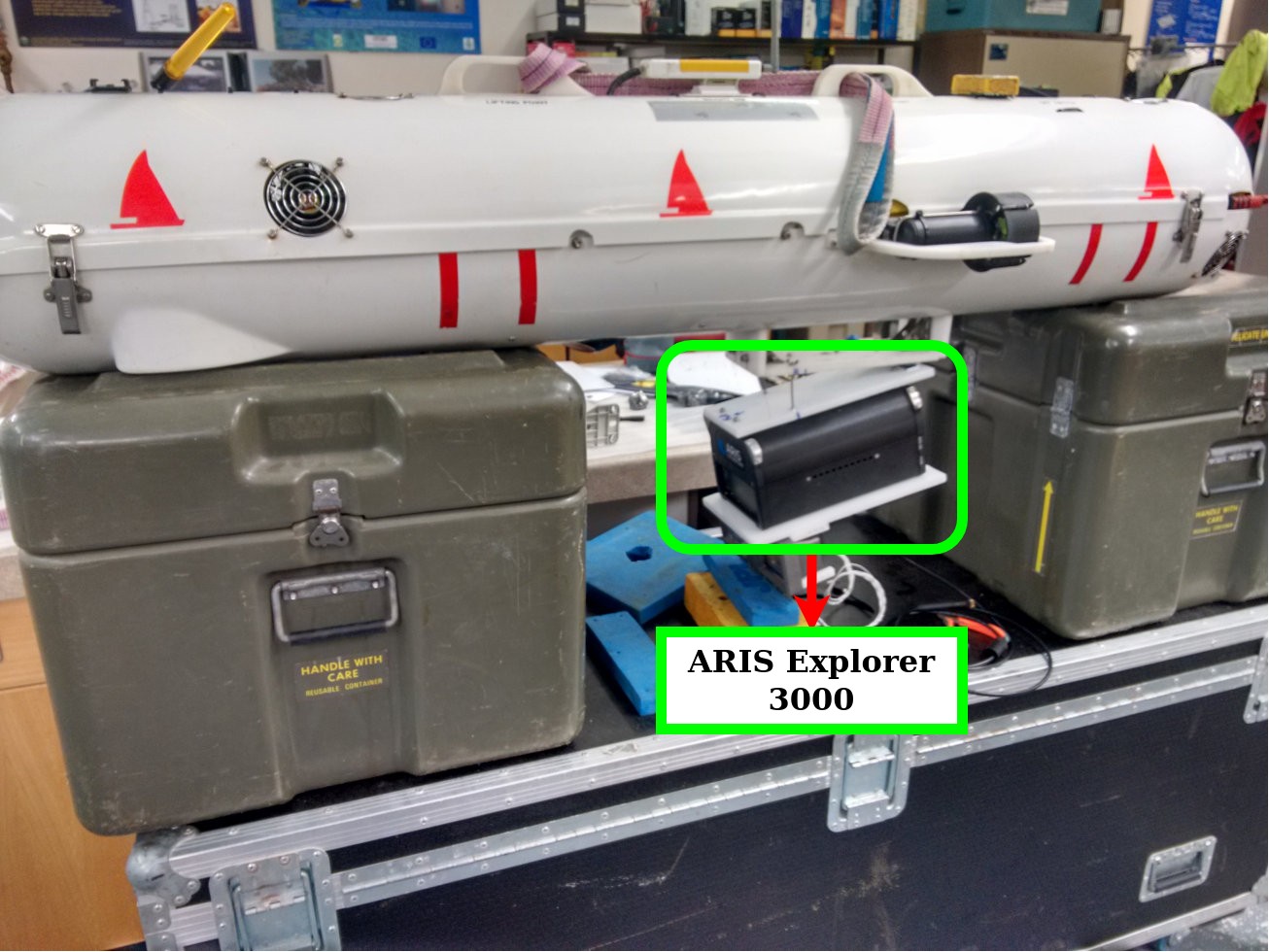}
    \caption{Nessie AUV with mounted FLS sensor (inside the bounding box)}
    \label{fig:nessie_fls}
\end{figure}

For collecting the data, a water tank of dimension $(W, \ H, \ D)\ = (3 \ \times \ 2 \times \ 4) \ m^{3}$ was used. In this water tank, an AUV was submerged with a sonar attached to its underside as shown in Figure \ref{fig:nessie_fls}. The difficulties of collecting marine debris data in a real world environment led to using an artificial water tank.  ARIS Explorer 3000 \cite{arisExplorer3K} forward looking sonar was used for collecting the data. Because of its high data frequency (almost 15 Hz), it acts like an acoustic camera. Also, given its high spatial resolution, each pixel can represent up to 2.3 millimeters of the environment. 

The above mentioned sonar has 128 acoustic beams and a field of view $30^\circ \ \times 15^\circ$ with spacing between beams being $0.25^\circ$. This sonar has a minimum distan	ce range of 70 cm and the maximum distance range varies with the sampling frequency. At a frequency of 1.8 MHz, the maximum range is around 5 meters. The spatial resolution of the sonar is 2.3 mm per pixel in close range and almost 10 cm per pixel at the far range \cite{valdenegro2019deep}. The sonar was mounted in a way to the AUV such that, it had a pitch angle between $15^{\circ}$ and $30^{\circ}$, so that the acoustic beam could insonify the bottom surface of water tank and enable clear observation of objects in the output sonar images. This assumes a flat world \cite{aykin2013forward}.

\begin{figure}[!t]
    \centering
    \includegraphics[scale=1.0, width=0.35\textwidth]{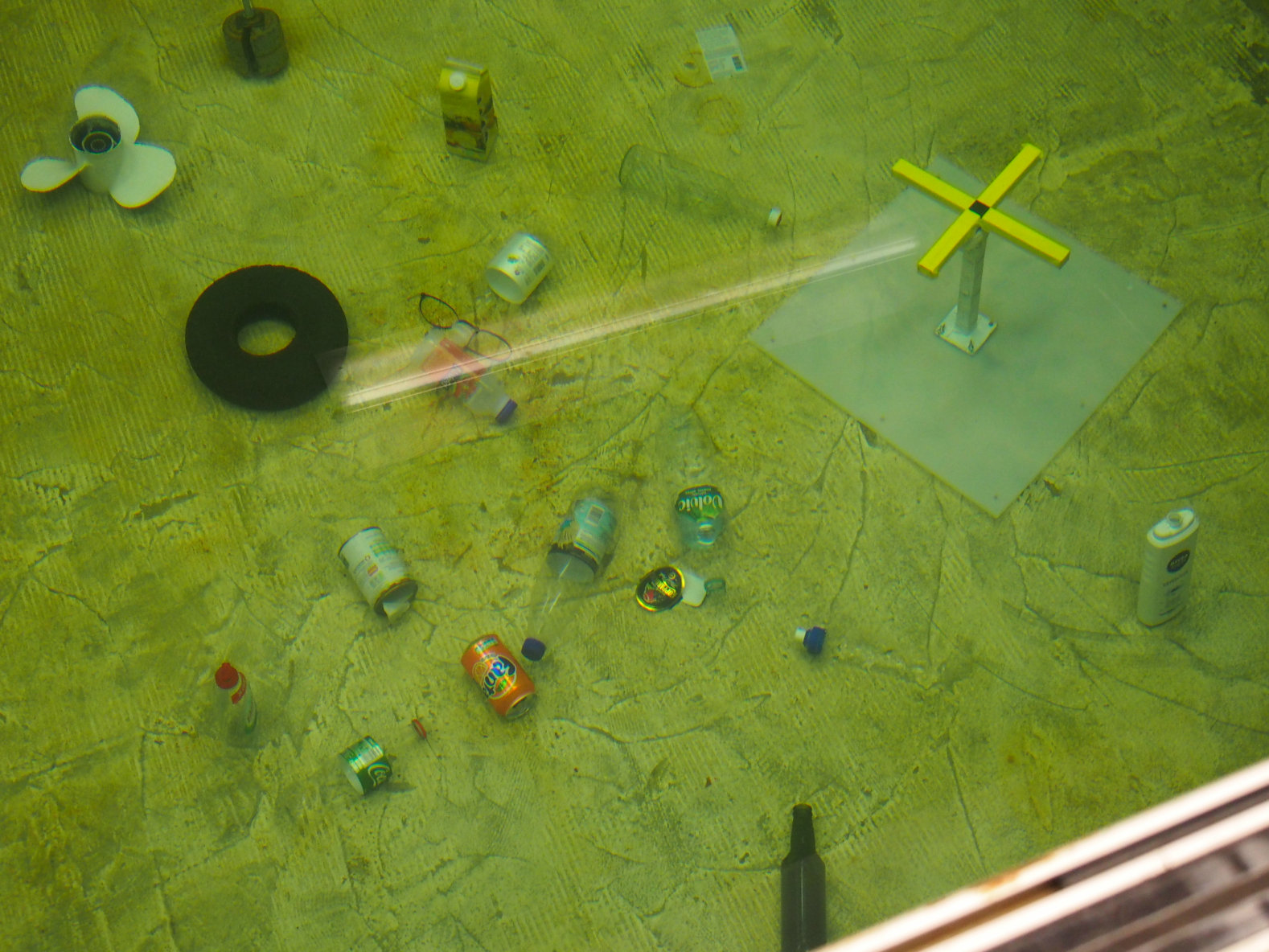}
    \caption{A sample scene of the water tank with objects kept inside}
    \label{fig:samples}
\end{figure}

The objects used for collecting the data were a combination of household and typical marine objects.  The household objects included materials like drink cartons, tin cans, drink cartons and plastic shampoo bottles. On the other hand, for marine objects, chain, hook, rubber tire, propeller, rubber tire and a valve mockup were used. Figure \ref{fig:samples} shows the water tank with the objects kept inside. This collection of objects was motivated by the fact that large portion of marine debris consists of discarded household objects \cite{valdenegro2019deep}. Along with these objects, the walls of the water tank were also considered. 

For this data collection, the sonar sensor was operated with a sampling frequency of 3.0 MHz and ARIScope application was used to set the parameters like gain etc. The captured data in the form of \texttt{.aris} files was projected into a polar field of view and saved as PNG image file. Out of all the images, only those were selected which contained image of at-least one object with clear visibility. Also, in order to avoid any temporal correlation between the frames, there was a time gap of atleast 5 frames between two selected images.

\begin{figure*}[ht]
\begin{center}
\subfloat[bottle]{
\includegraphics[width=.32\columnwidth]{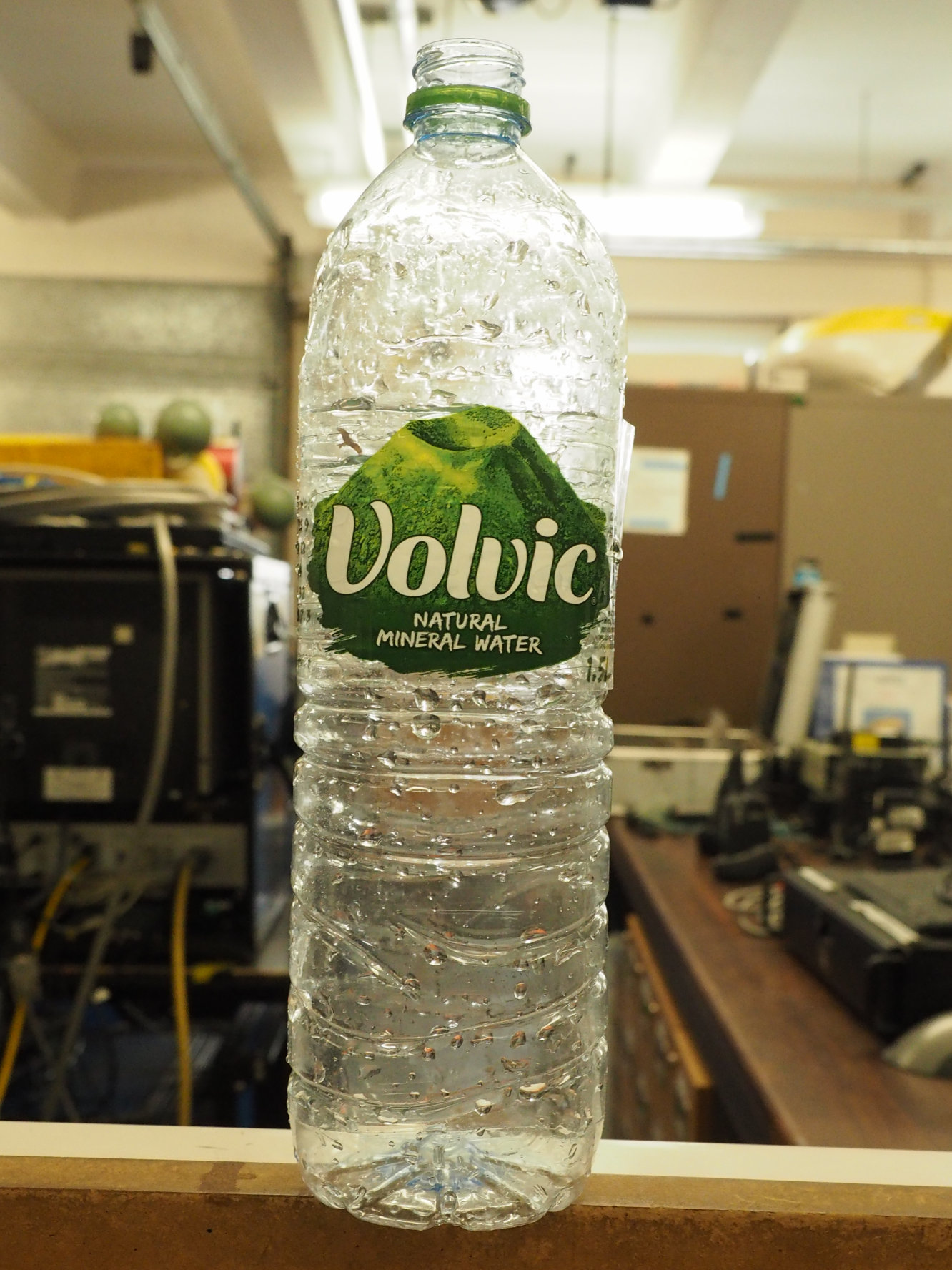}
}
\subfloat[can]{
\includegraphics[width=.32\columnwidth]{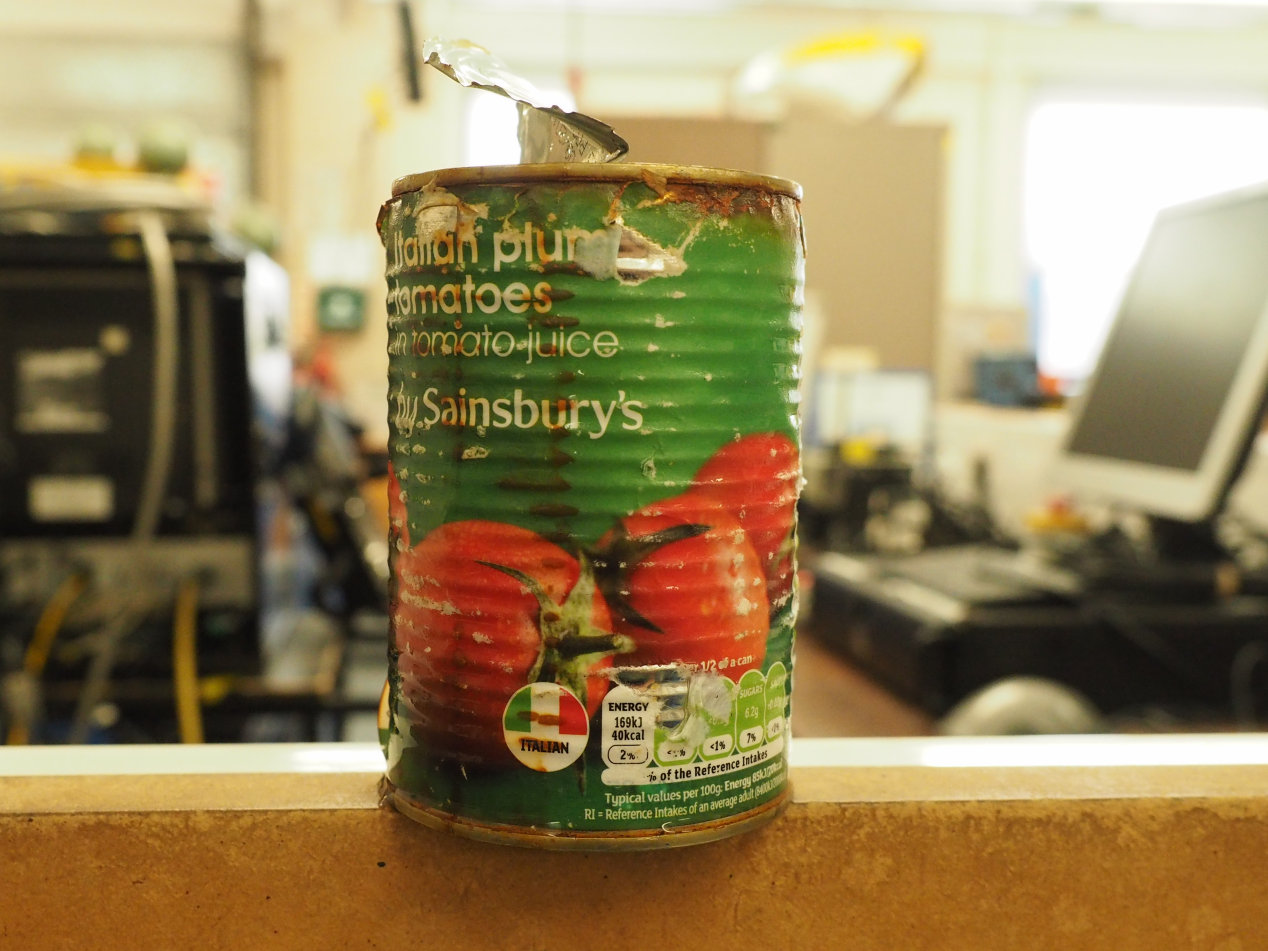}
}
\subfloat[chain]{
\includegraphics[width=.32\columnwidth]{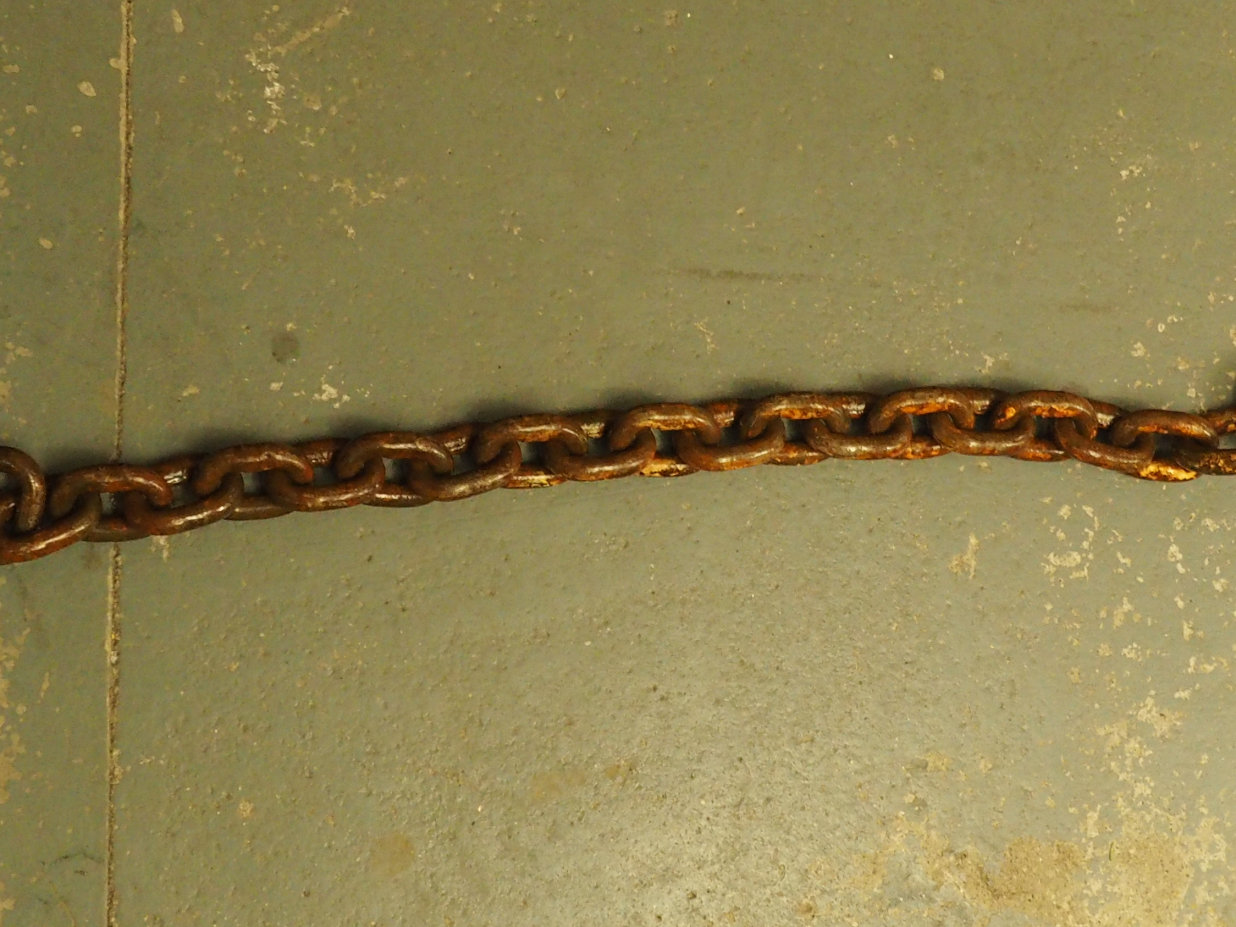}
}
\subfloat[drink-carton]{
\includegraphics[width=.32\columnwidth]{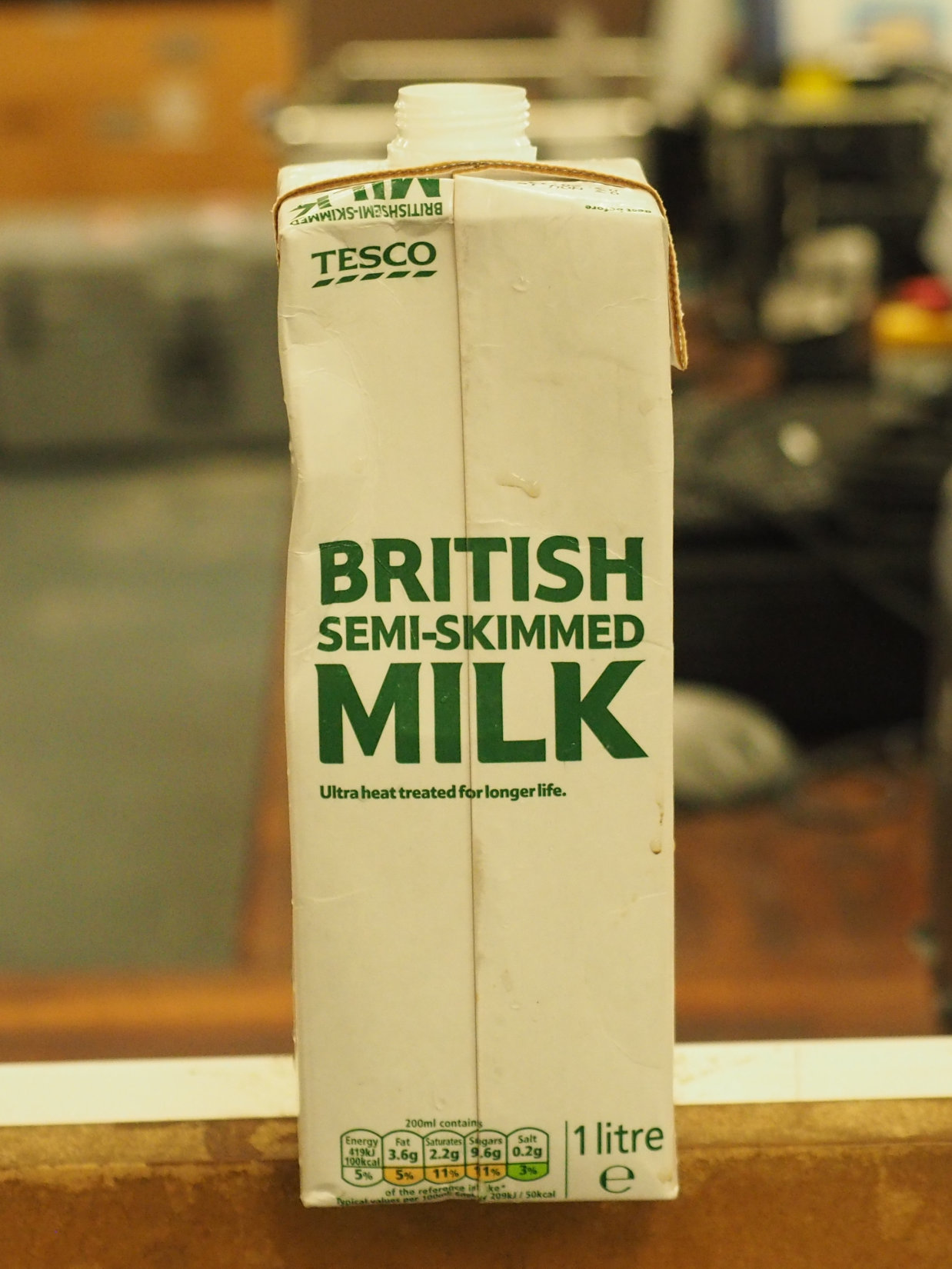}
}
\subfloat[hook]{
\includegraphics[width=0.32\columnwidth]{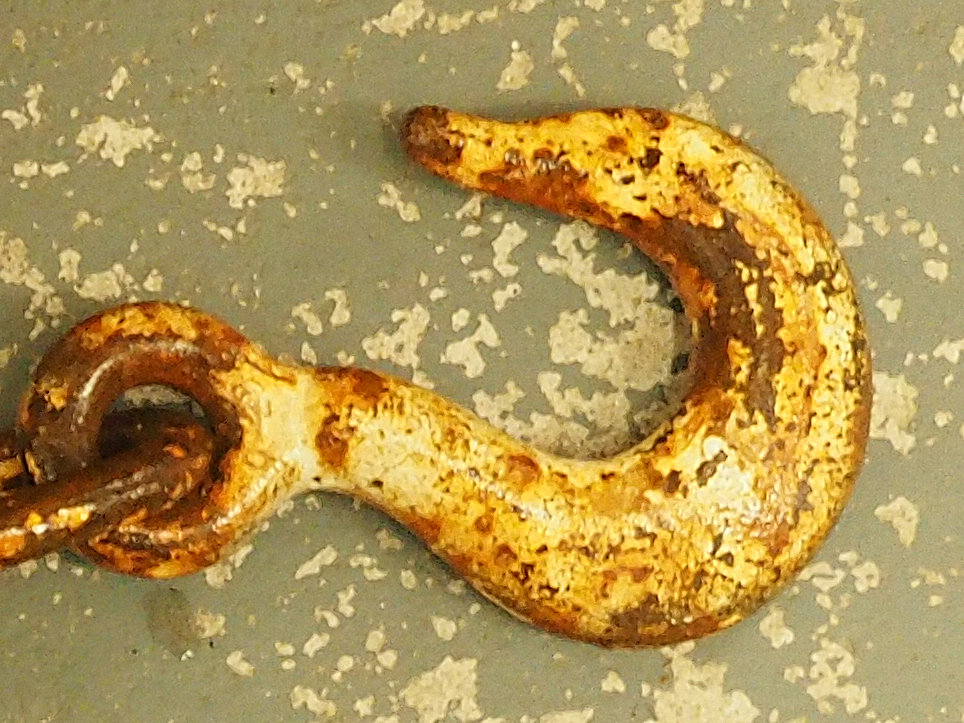}
}
\subfloat[propeller]{
\includegraphics[width=0.32\columnwidth]{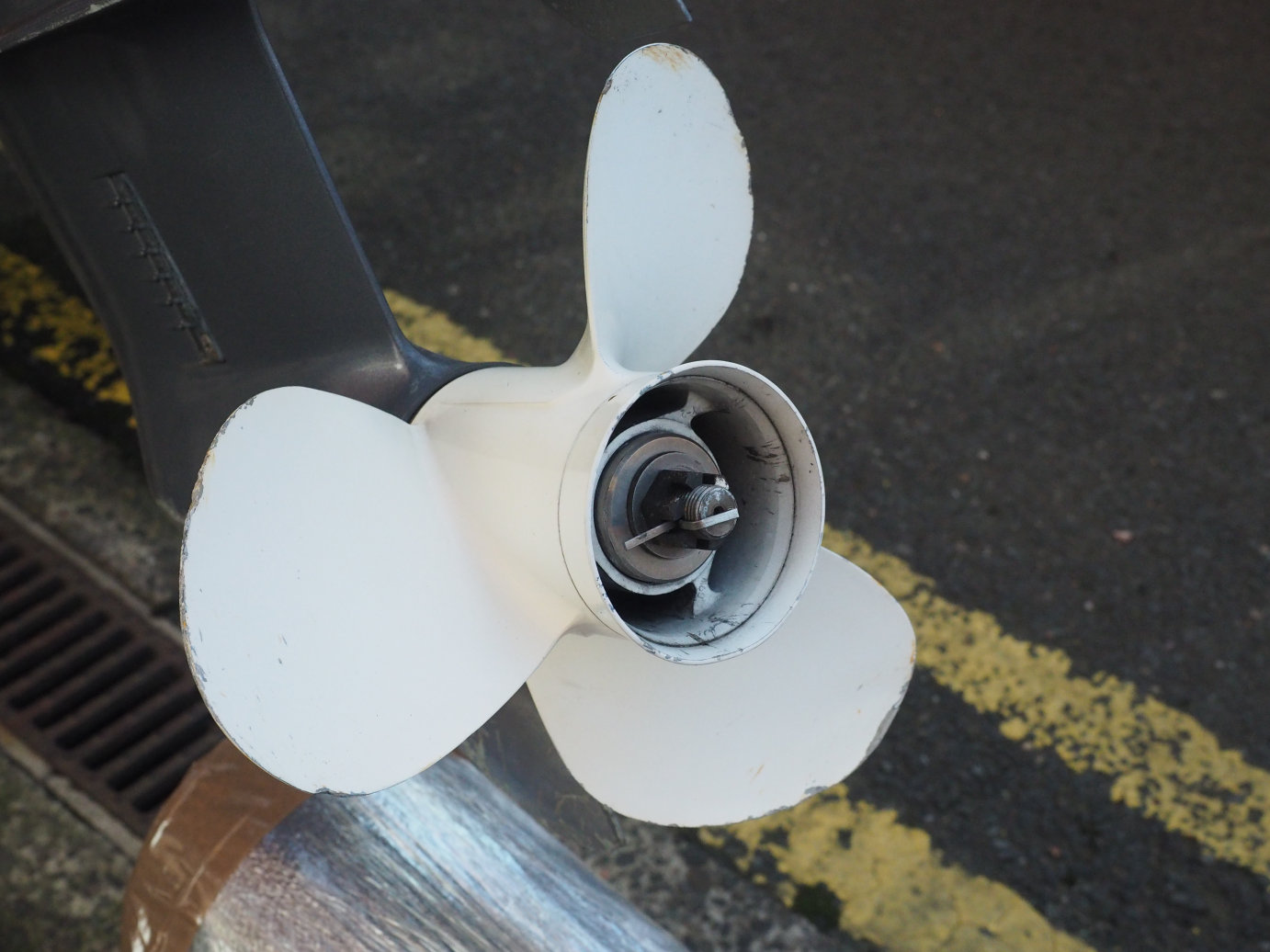}
}

\subfloat[shampoo-bottle]{
\includegraphics[width=0.32\columnwidth]{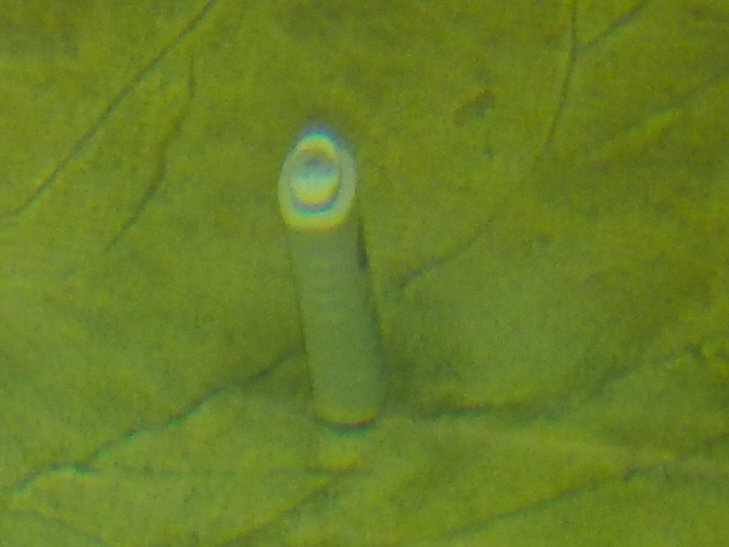}
}
\subfloat[standing-bottle]{
\includegraphics[width=0.32\columnwidth]{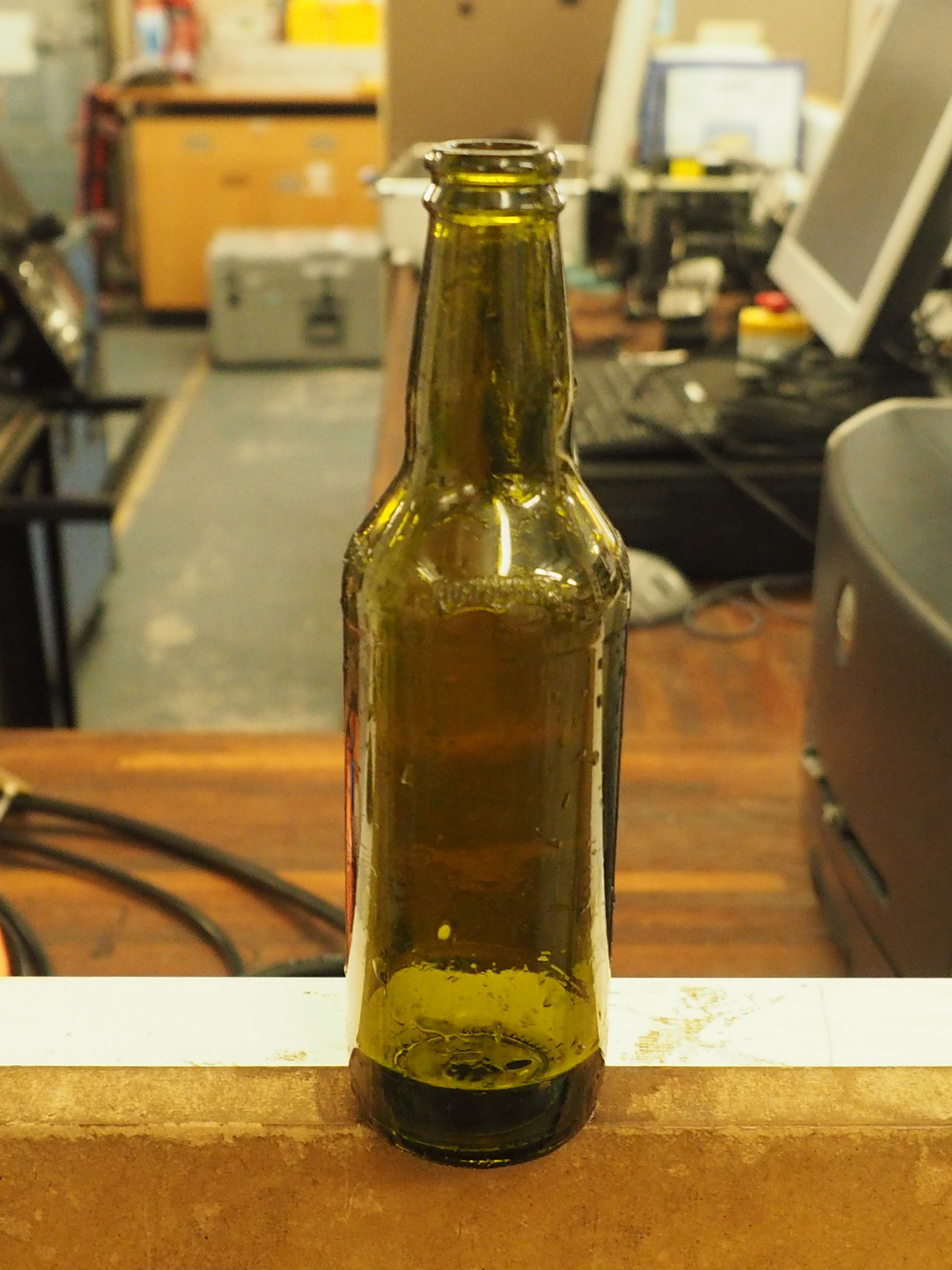}
}
\subfloat[tire]{
\includegraphics[width=0.32\columnwidth]{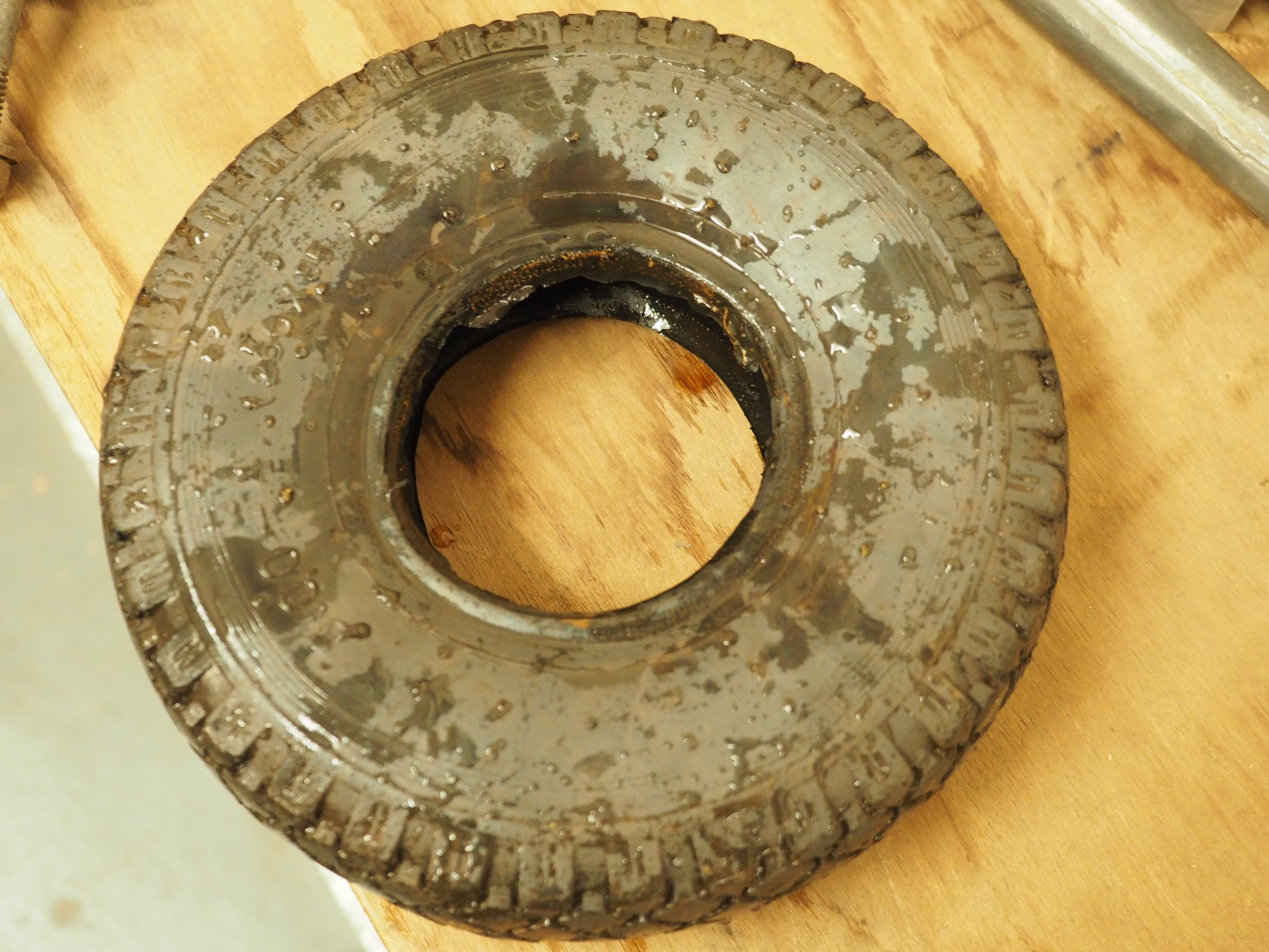}
}
\subfloat[valve]{
\includegraphics[width=0.32\columnwidth]{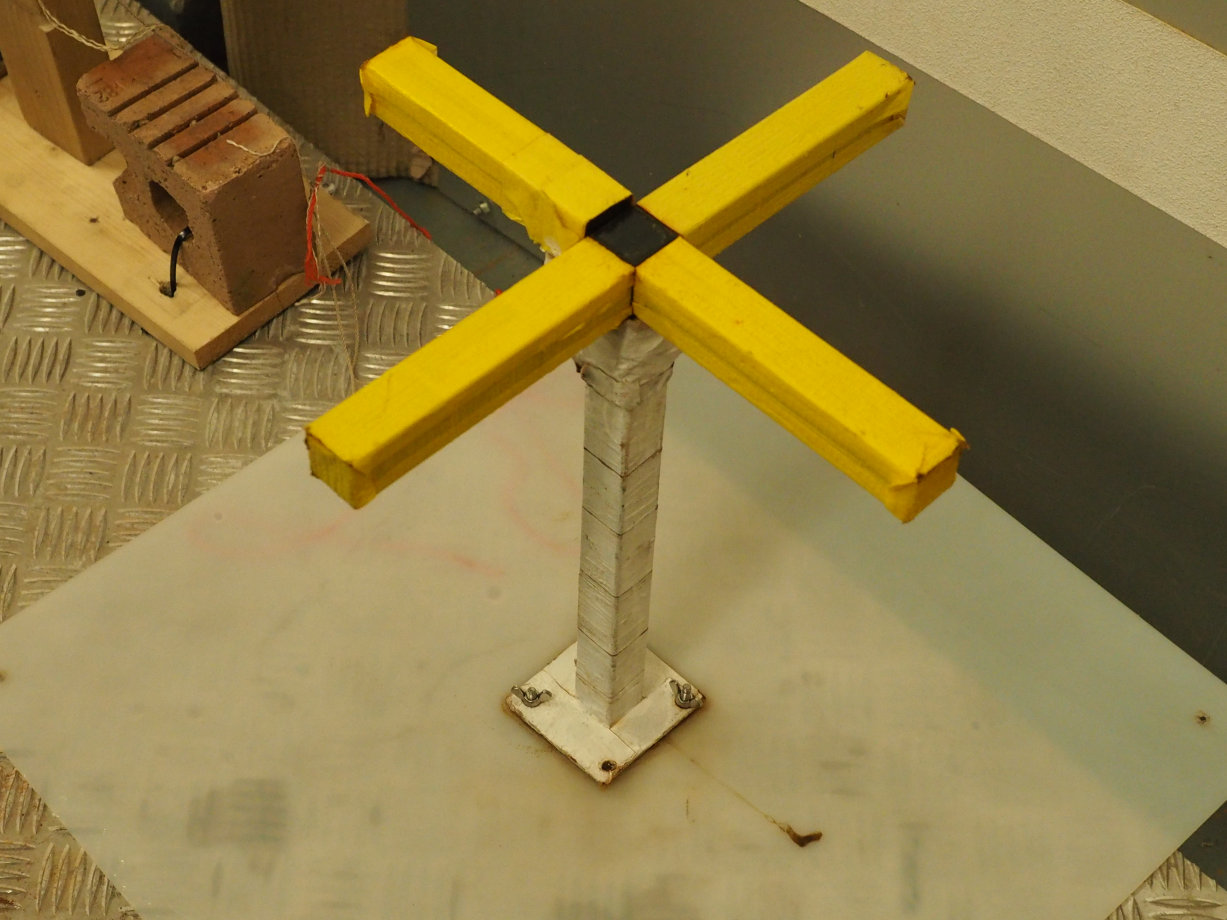}
}
\subfloat[wall]{
\includegraphics[width=0.32\columnwidth]{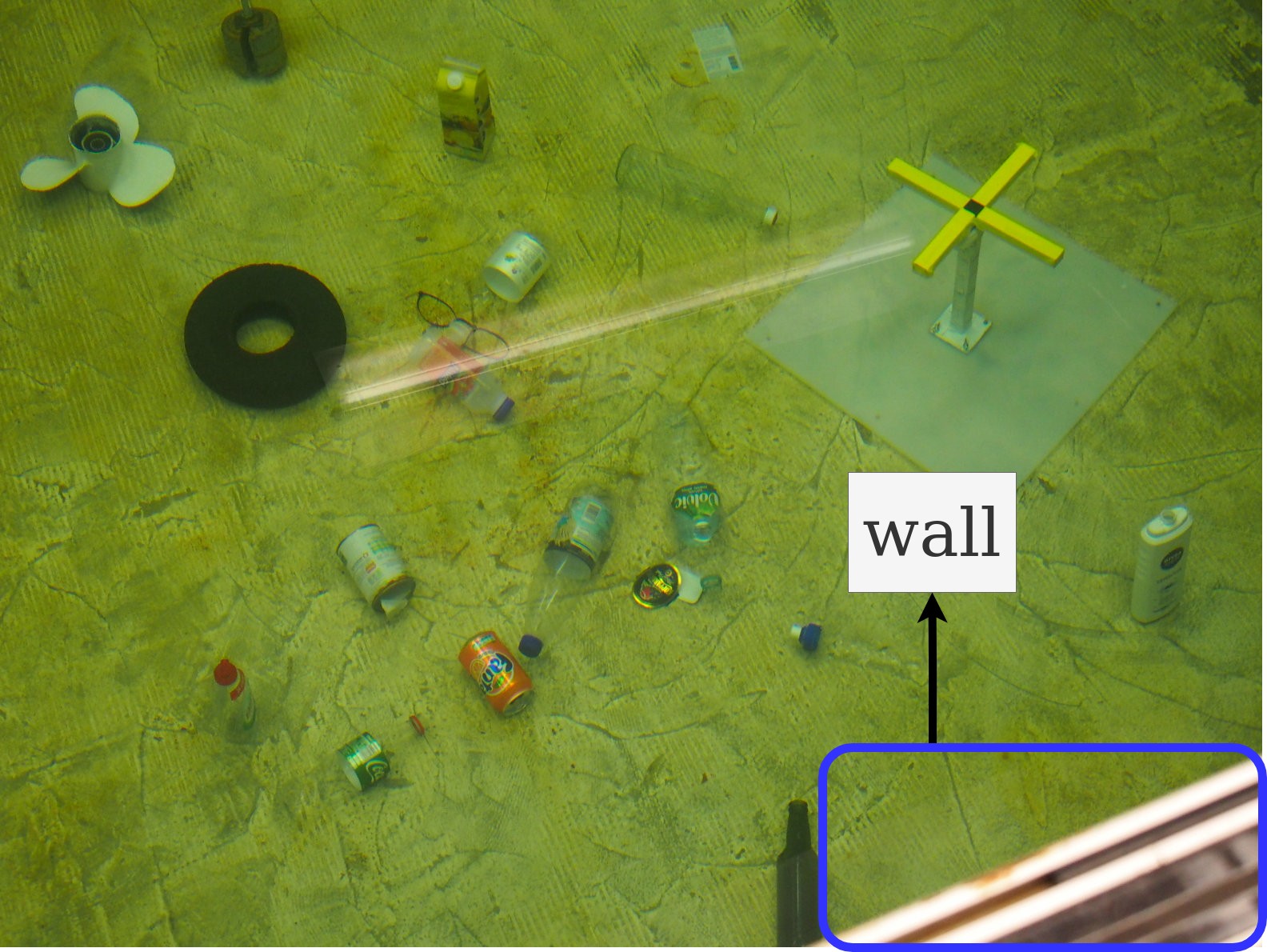}
}

\caption{Types of objects used to create the dataset.}
\label{fig:data}
\end{center}

\end{figure*}

The AUV was teleoperated in the water tank to collect the data. It moved at a velocity of 0.1 m/s inside the water tank. The robot was moved around in the tank on a quarter of a circle path at a speed of around 0.1 m/s. Given the limited space to move inside the water tank, the sensor could not capture all the data for each object from all perspectives. Figure \ref{fig:data} shows the objects that were used to collect the data. For image of wall, boundary of the tank has been highlighted

\begin{figure*}
\begin{center}
\subfloat[FLS image 1]{
\includegraphics[width=0.4557\columnwidth]{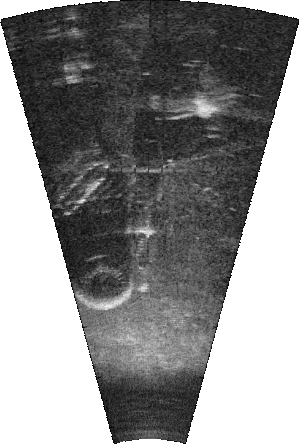}
}    
\subfloat[FLS image 2]{
\includegraphics[width=0.465\columnwidth]{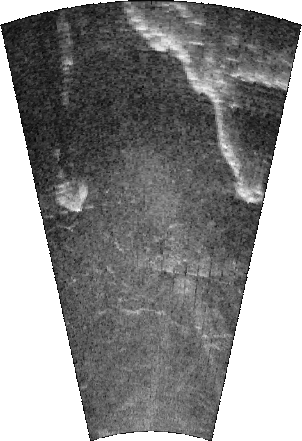}
}  
\subfloat[FLS image 3]{
\includegraphics[width=0.465\columnwidth]{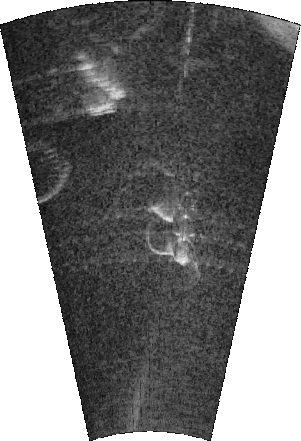}
}
\subfloat[FLS image 4]{
\includegraphics[width=0.465\columnwidth]{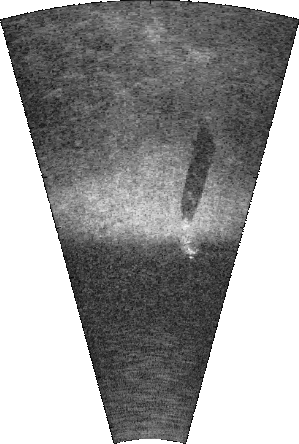}
}  
\\
\subfloat[FLS image 1 labels]{
\includegraphics[width=0.4557\columnwidth]{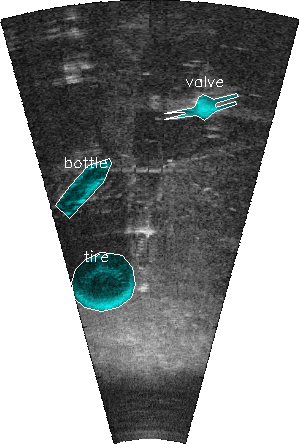}
}    
\subfloat[FLS image 1 labels]{
\includegraphics[width=0.465\columnwidth]{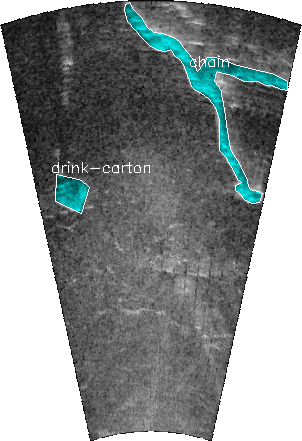}
}  
\subfloat[FLS image 3 labels]{
\includegraphics[width=0.465\columnwidth]{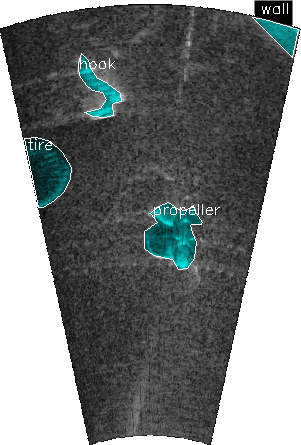}
}
\subfloat[FLS image 4 labels]{
\includegraphics[width=0.465\columnwidth]{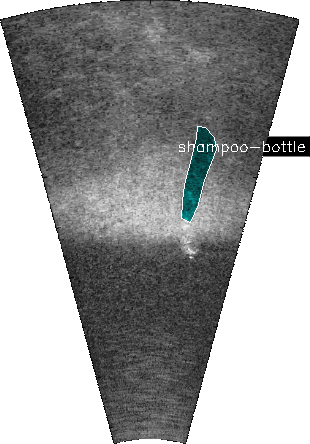}
}  
\end{center}
\caption{Sample sonar images from FLS sensor (upper row) and corresponding segmentation masks and class labels (bottom row).}
\label{fig:labels}
\end{figure*}

\subsection{Object Classes and Annotations}

\begin{table*}[t]
\begin{center}
 \begin{tabular}{lll} 
 \toprule
 Class ID & Object & Description \\ [0.5ex] 
 \midrule
 0 & background & Everthing except objects\\ 
 1 & bottle &  Glass and plastic bottles lying horizontally (Figure \ref{fig:data}(a))\\
 2 & can & Several metallic cans  (Figure \ref{fig:data}(b))\\
 3 & chain & One meter long chain with small chain links (Figure \ref{fig:data}(c)) \\
 4 & drink-carton & Horizontally lying milk/juice cartons (Figure \ref{fig:data}(d))\\
 5 & hook & Small metallic hook (Figure \ref{fig:data}(e)) \\
 6 & propeller & A small mettalic ship propeller (Figure \ref{fig:data}(f)) \\
 7 & shampoo-bottle & A vertical shampoo bottle (Figure \ref{fig:data}(g)) \\
 8 & standing-bottle & A standing beer bottle made of glass (Figure \ref{fig:data}(h))\\
 9 & tire & Horizontally lying small rubber tire (Figure \ref{fig:data}(i)) \\
 10 & valve & A mock-up metal valve (Figure \ref{fig:data}(j))\\
 11 & wall & Boundary of the water tank (highlighted part in Figure \ref{fig:data}(k)) \\ [1ex] 
 \toprule
\end{tabular}
\caption{Selection of semantic classes available in our dataset}
\label{tab:dataTab}
\end{center}
\end{table*}

The dataset consists of 1868 images with total number of object classes being 11 (excluding background). The details of each object with the class number are shown in Table \ref{tab:dataTab}. The images were standardized to a size of $480 \times 320$ pixels and then they were labeled with per-pixel annotations. 

\begin{figure}[h]
\begin{center}
\subfloat[Glass Bottle]{
\includegraphics[width=.25\columnwidth]{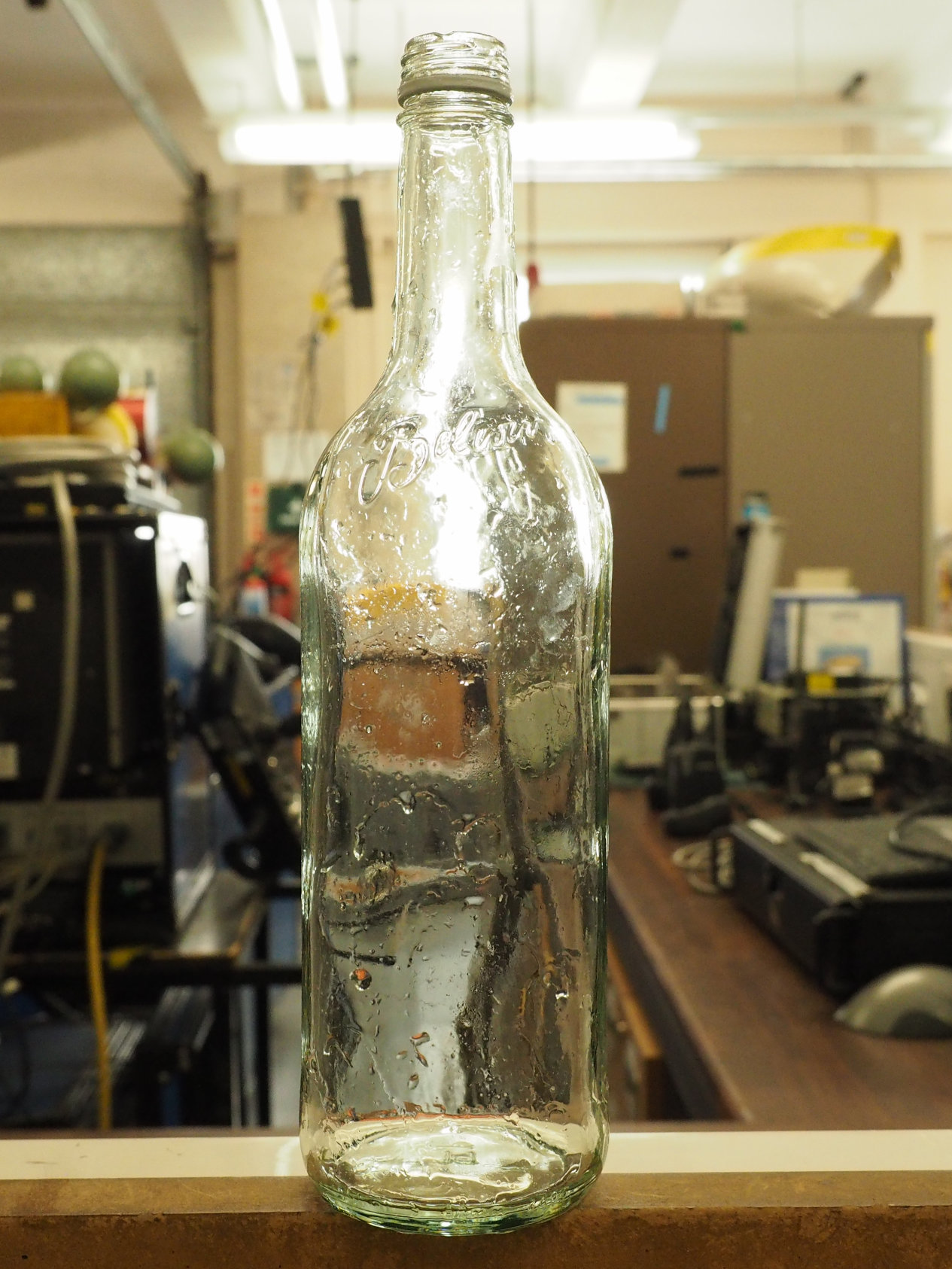}
}
\subfloat[Plastic Bottle]{
\includegraphics[width=.25\columnwidth]{Images/plastic-bottle.jpg}
}
\caption{Glass and plastic bottles that belong to the same class (bottle)}
\label{fig:bottle}
\end{center}
\end{figure}

As can be seen in Table \ref{tab:dataTab}, there are 3 different classes for bottles. All the bottles that lie on horizontally on the floor of the tank are grouped together in one class (bottle class). Figure \ref{fig:bottle} shows two bottles, a glass bottle and a plastic bottle which belong to the same class (bottle) as these lie horizontally in the tank. The beer bottle standing on the floor of water tank was assigned a separate class (standing-bottle) as it looked different from other bottles in the sonar image.  Also, the standing shampoo bottle appeared differently and hence assigned a different class. The walls of the tank were also labelled as a separate class different from background.

The images were labeled for semantic segmentation using LabelMe \cite{russell2008labelme}. Absence of color information and presence of significant noise posed a challenge while labeling sonar data. Not all objects were clearly visible in all images and most of the objects did not produce a shadow. The appearance of shadow of objects varied with their size and the perspective at which there images were taken. The main aim of the annotator was to label the highlight of all objects, and for difficult objects, to label the shadow, as it would be the most significant feature for detection.

\begin{figure}
\centering
\subfloat[Original Image]{
\includegraphics[width=0.40\columnwidth]{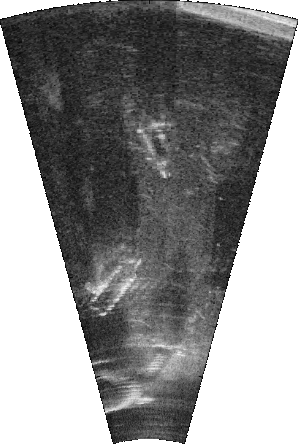}
}
\subfloat[Background pixels]{
\includegraphics[width=0.40\columnwidth]{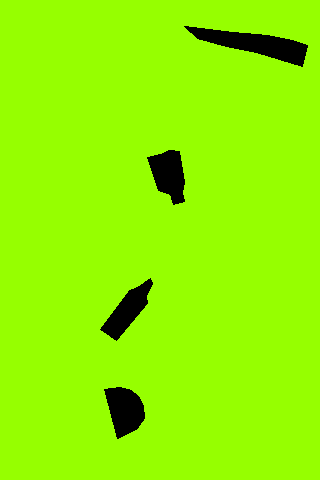}
}
\caption{Displaying background pixels in the image. Colored pixels in the image on right represent the background.}
\label{fig:bg}
\end{figure}
Everything other than the objects has been considered as background in the dataset. As can be seen from Figure \ref{fig:bg}, the background consists not only all the pixels lying in the portion of image which is outside the field of view of sonar (black portion in Figure \ref{fig:bg} (a)) but also occupies a major portion of the mapped part which has variable intensity values and sometimes equal to those of objects being labeled. Keeping this is mind, the object boundaries were labeled very carefully so as to precisely capture the shape of the object. As can be seen from Figure \ref{fig:labels} (e), shape of the tire makes it easily detectable in the sonar image.   

\subsection{Train, Test and Validation Splits}
Out of 1868 images, 1000 were selected for training, 251 for validation and 617 for testing. 

The data was divided randomly among the 3 sets for a few iterations until the distribution of pixel and object count for each class across the 3 splits matched. This ensured that no bias was there while dividing the data between training, testing and validation sets. 

\subsection{Dataset Statistics}

In this section we present some statistics of the dataset we captured.

\begin{figure}[h]
    \centering
    \includegraphics[scale=1.0,width=0.45\textwidth]{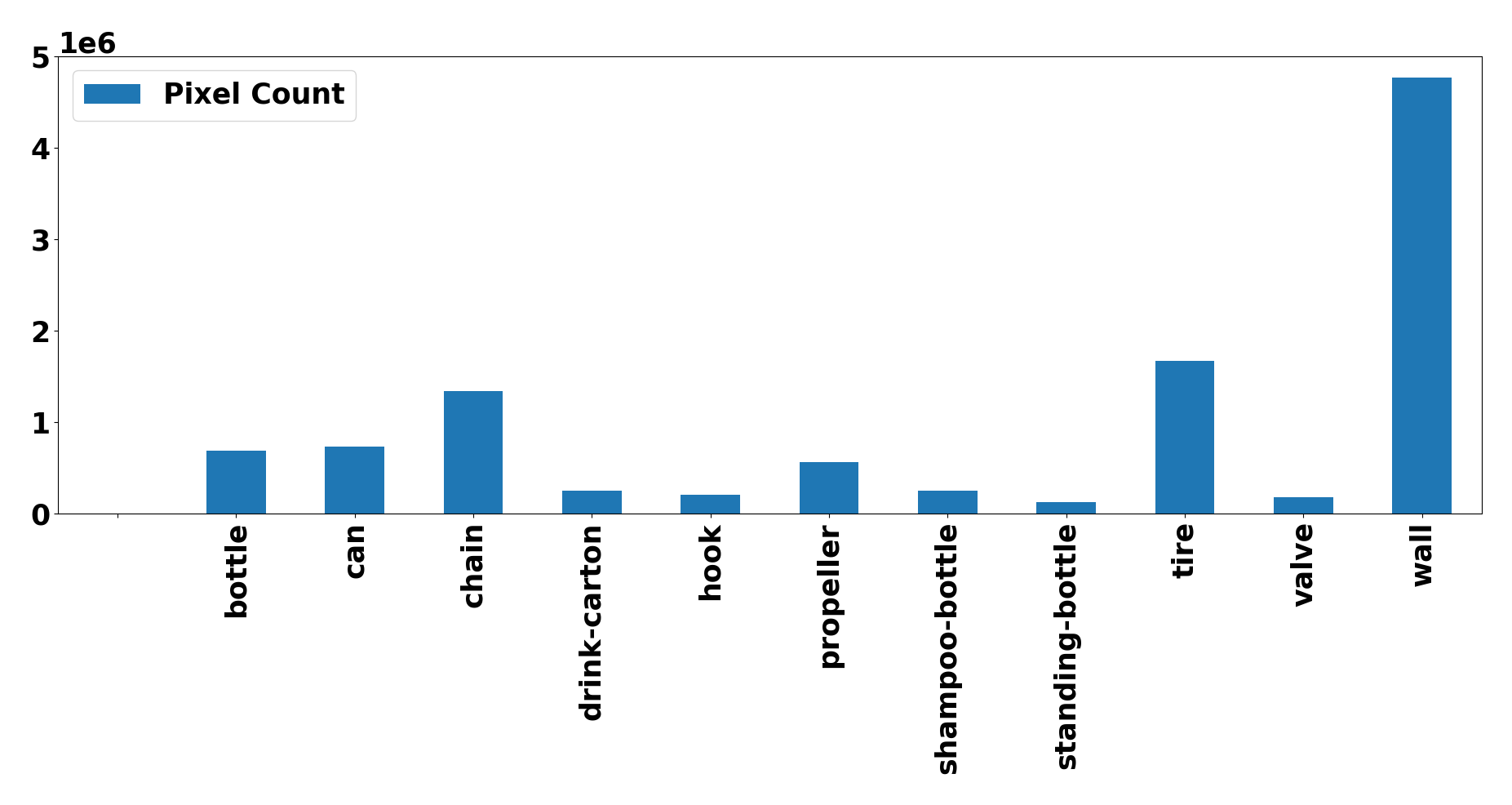}
    \caption{Total number of pixels for each class except background}
    \label{fig:pixHist}
\end{figure}
Figure \ref{fig:pixHist} shows the total number of pixels across the whole dataset which belong to each object (except background). 


\begin{figure}[h]
    \centering
    \includegraphics[scale=1.0,width=0.45\textwidth]{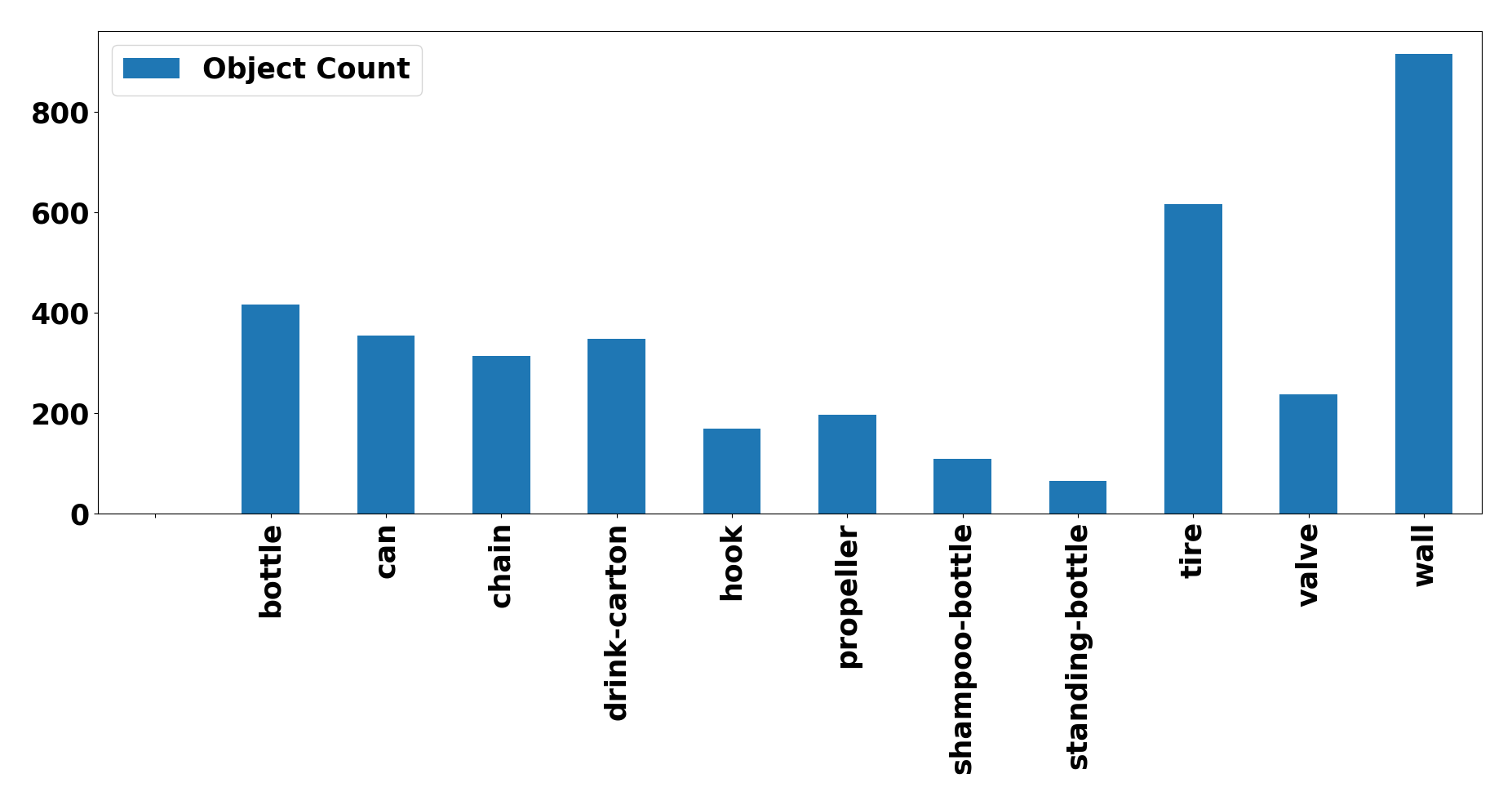}
    \caption{Total count of occurrence of each object class except background}
    \label{fig:countHist}
\end{figure}

Since total pixel counts not only depends on the occurrence of the objects in the image but also on their size, total pixel count may be biased towards objects that have large size than others. Figure \ref{fig:countHist} shows the count of appearance of each object in the dataset. Background is not taken into consideration in both Figures \ref{fig:pixHist} and \ref{fig:countHist} but it is given a separate class for segmentation. Watertank walls have the highest pixel count, since they appear in many images and produce a strong acoustic response, while other object classes have similar balance in number of pixels, except for small objects like bottles and the mockup valve.

\begin{figure}
    \centering
    \includegraphics[scale=1.0, width = 0.5\textwidth]{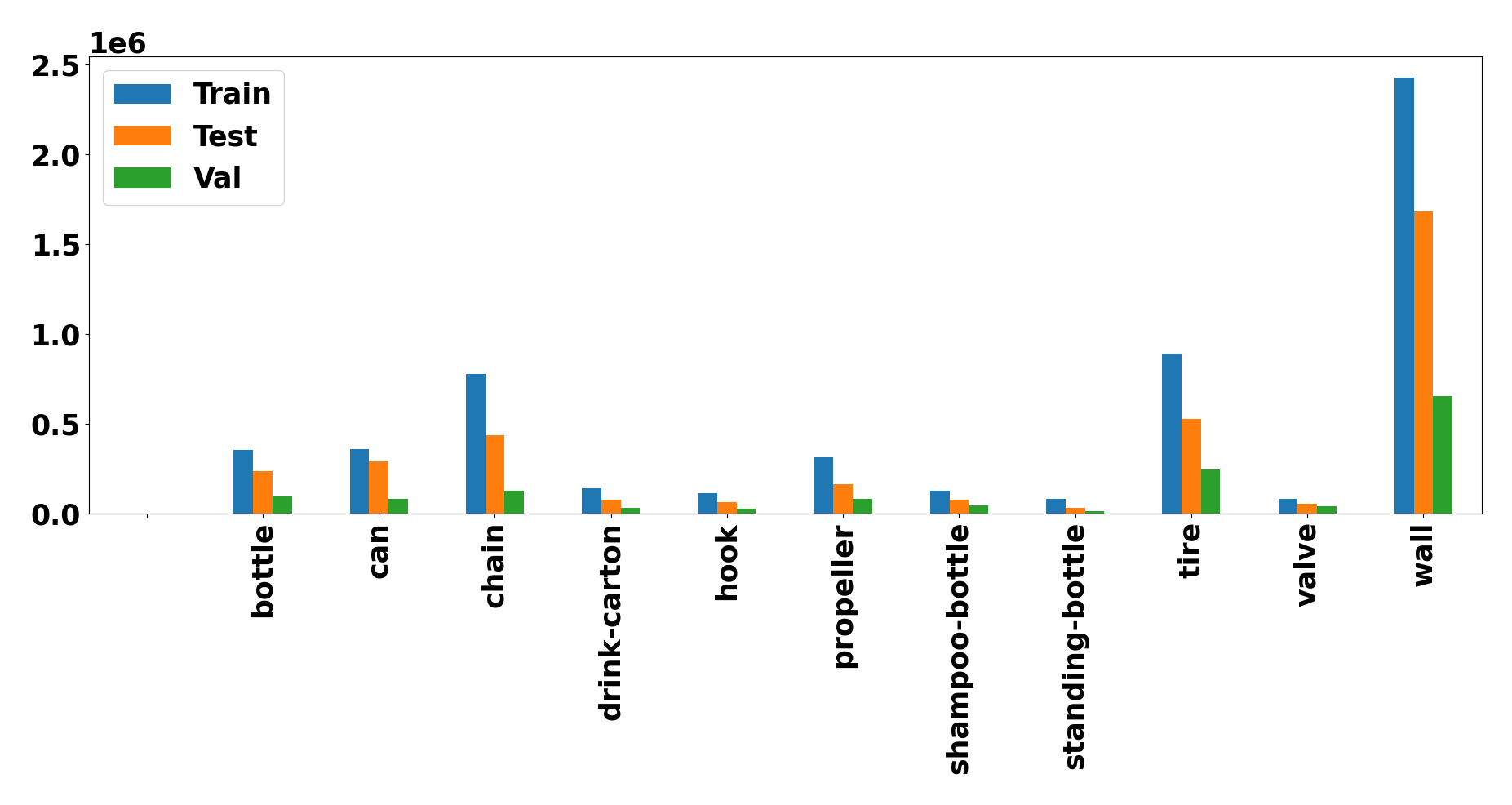}
    \caption{Total pixel count for each class in training, testing and validation sets}
    \label{fig:split}
\end{figure}

\begin{figure}
    \centering
    \includegraphics[scale=1.0, width = 0.5\textwidth]{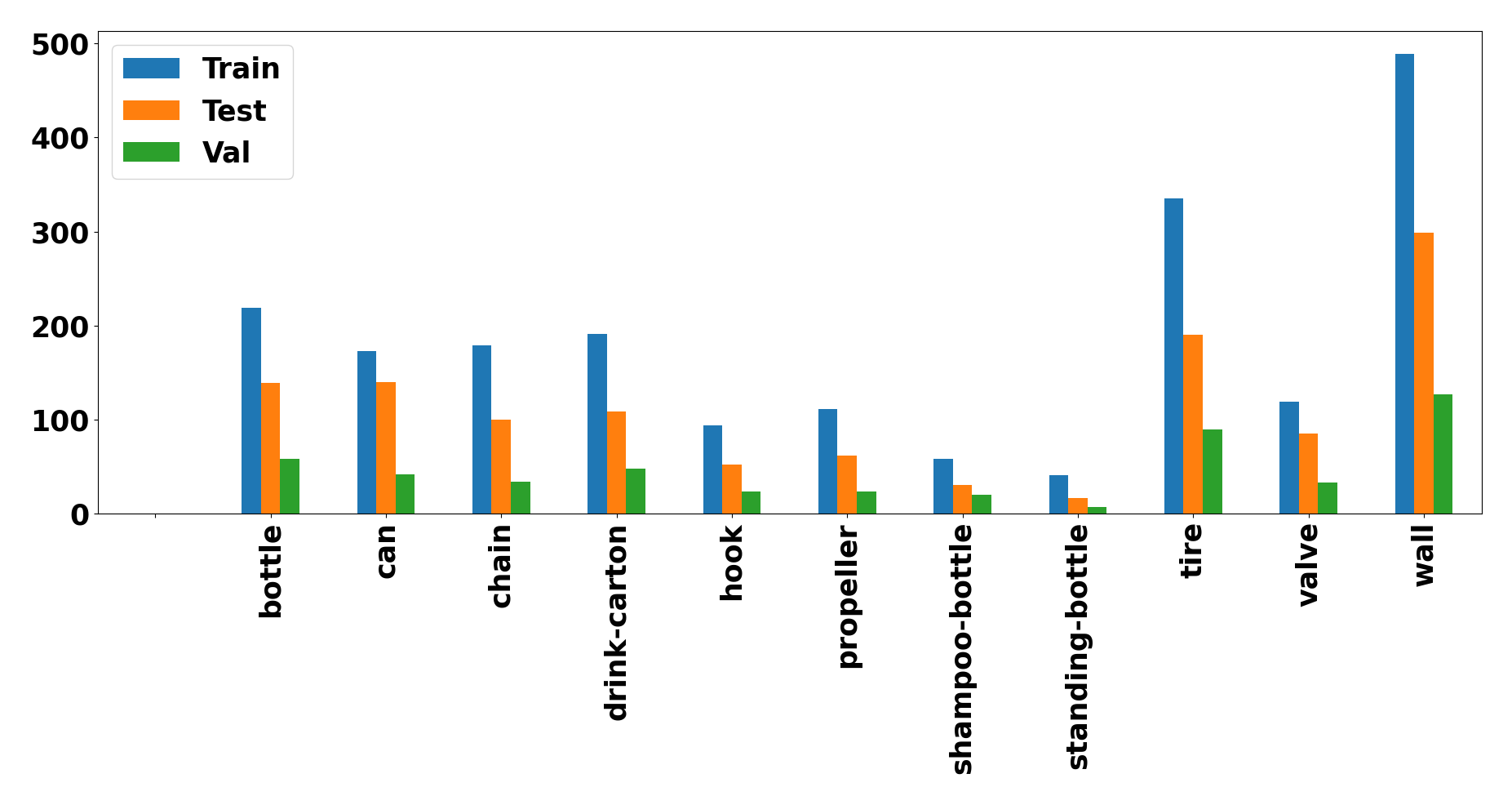}
    \caption{Total object count for each class in training, testing and validation sets}
    \label{fig:split_object}
\end{figure}

Figures \ref{fig:split} and \ref{fig:split_object} show the total pixel count and object count for each class across training, testing and validation splits. As can be seen, the distribution remains similar in all the splits.

\section{Semantic Segmentation Baselines}
This section describes the details of semantic segmentation baselines on this dataset.

\subsection{Segmentation Models}
For generating the baseline semantic segmentation results on the presented dataset, four main segmentation architectures were used, namely Unet \cite{ronneberger2015u}, LinkNet \cite{chaurasia2017linknet}, PSPNet \cite{zhao2017pyramid} and DeepLabV3 \cite{chen2017rethinking}. The architectures are based on encoder-decoder models and were selected as they are state of the art in many datasets and tasks, and represent a good variety of segmentation architectures.

UNet, built upon Fully Connected Networks (FCNs) is used popularly for biomedical image segmentation. The encoder or contracting part comprising of convolutional with $3 \times 3$ filters and pooling layers is very similar to that of FCN. In the decoder part, for precise localization of features obtained from the encoder, the feature maps obtained at various stages of contracting part are concatenated with the corresponding upsampled feature maps in the decoder.

LinkNet is another encoder-decoder based segmentation architecture that contains fewer parameters than any other architecture used in this paper. The encoder part of LinkNet consists of blocks of convolution, batch normalization, relu activation and skip connection. The decoder uses full convolution proposed in \cite{long2015fully}. 

Pyramid Scene Parsing Network (PSPNet) uses global contextual features in combination with local context features obtained from repeated convolutional operations. It uses a pyramidal pooling module to aggregate the regional context information into an FCN based segmentation architecture. 

DeepLabV3  uses atrous convolution to prevent filter decimation and adjust the field of view of filters. It uses modules which employ atrous convolution operations with multiple feature resolutions in cascaded or parallel manner for capturing features at multiple scales. It augments the atrous spatial pyramidal pooling used in DeepLabV2 by adding image-level features for encoding the global context.

For each of these segmentation architectures, a variety of encoders are used. The encoders used are ResNet(18, 34, 50, 101, 152) \cite{resnet}, VGG (16, 19) \cite{vgg}, EfficientNet (b0, b1 and b2) \cite{EfficientNetRM}, InceptionNetV3 \cite{inception}, and InceptionResNetV2 \cite{szegedy2016inceptionv4}. These encoders have a high variation in there architectures and depths. The results generated by using these encoders in the above mentioned segmentation architectures give an insight to how each combination performs on a data like this which is very different from conventional segmentation datasets as they have a lot more distinct features like colors, etc.

\subsection{Metrics}
To analyze the performance of various semantic segmentation architectures on this dataset, intersection over union (IoU) metric has been used \cite{cordts2016cityscapes}. The IoU values are obtained on the training, testing and validation sets. For the training and validation sets, the IoU value calculated on the last training epoch are used. 

Since mean IoU ($mIoU$) over all classes gets affected by the presence of large objects or background pixels present in the image (which in the presented dataset occupies a large portion of the image), it may not indicate the performance of segmentation algorithms properly for all the classes as many objects are smaller in size (e.g. cans are extremely small as compared to wall). To solve this problem, along with $mIou$, per-class mean IoU ($mIoU_{i}$) across all the images for each split is also computed for each segmentation algorithm. Here $i$ refers to class index.

\subsection{Training Details}
As the data consisted of multiple classes, categorical cross entropy loss was used for segmentation. For optimization, ADAM optimizer \cite{kingma2014adam} was used with learning rate $0.001$ and the exponential decays for 1st and 2nd moment estimates as 0.9 and 0.999.

The batch size was chosen to be 16 and the maximum number of epochs for training the models was set to 30. In order to avoid over-fitting, early stopping was used with the criteria of stopping training when loss value did not change for 3 consecutive epochs. Since the dataset consists of grayscale images and not colored like ImageNet or CityScapes, the models were trained from scratch without using any pre-trained weights.

For segmentation using DeepLabV3, only Resnet50, VGG16 and VGG19 encoders have been used. The cascaded version of DeepLabV3 has been used in this case with these encoders and the output stride (O.S.) was kept to be 8.  

\subsection{Baseline Results}

\begin{table}[h!]
\begin{center}
 \begin{tabular}{llll} 
 \toprule
  Encoder & $\text{mIoU}_\text{train}$ & $\text{mIoU}_\text{val}$ & $\text{mIoU}_\text{test}$ \\ [0.5ex] 
 \midrule
 RN18 & 0.6572 & 0.5912 & 0.6174 \\ 
 RN34 & 0.7732 & 0.7169 & 0.7481 \\
 RN50 & 0.7881 & 0.6967 & 0.7041 \\
 RN101 & 0.5576 & 0.5002 & 0.5165 \\
 RN152 & 0.5710 & 0.5541 & 0.5562 \\
 Vgg16 & 0.5604 & 0.5422 & 0.5374 \\
 Vgg19 & 0.5133 & 0.4737 & 0.4744 \\
 ENb0 & 0.6296 & 0.5699 & 0.5639 \\
 ENb1 & 0.6763 & 0.6684 & 0.6763 \\
 ENb2 & 0.5786 & 0.5128 & 0.5264 \\
 IncepV3 & 0.6213 & 0.5962 & 0.5956 \\
 IncepRNV2 & 0.7554 & 0.7130 & 0.7245 \\
 \toprule
\end{tabular}
\caption{Segmentation Results using Unet}
\label{tab:Unet}
\end{center}
\end{table}

\begin{table}[h!]
\begin{center}
 \begin{tabular}{llll} 
 \toprule
  Encoder & $\text{mIoU}_\text{train}$ & $\text{mIoU}_\text{val}$ & $\text{mIoU}_\text{test}$ \\ [0.5ex] 
 \midrule
 RN18 & 0.6934 & 0.6876 & 0.6923 \\ 
 RN34 & 0.7259 & 0.7039 & 0.7192 \\
 RN50 & 0.7955 & 0.6644 & 0.6697 \\
 RN101 & 0.6770 & 0.6053 & 0.6188 \\
 RN152 & 0.6489 & 0.5654 & 0.556 \\
 Vgg16 & 0.6053 & 0.5472 & 0.5458 \\
 Vgg19 & 0.4916 & 0.4298 & 0.4296 \\
 ENb0 & 0.7679 & 0.6826 & 0.7131 \\
 ENb1 & 0.6921 & 0.62464 & 0.6494 \\
 ENb2 & 0.7218 & 0.6703 & 0.6782 \\
 IncepV3 & 0.7172 & 0.6833 & 0.6948 \\
 IncepRNV2 & 0.6864 & 0.6184 & 0.6219 \\
 \toprule
\end{tabular}
\caption{Segmentation Results using LinkNet}
\label{tab:LinkNet}
\end{center}
\end{table}

\begin{table}[h!]
\begin{center}
 \begin{tabular}{llll} 
 \toprule
 Encoder & $\text{mIoU}_\text{train}$ & $\text{mIoU}_\text{val}$ & $\text{mIoU}_\text{test}$ \\ [0.5ex] 
 \midrule
 RN50 & 0.6771 & 0.6570 & 0.6678 \\
 Vgg16 & 0.6328 & 0.6087 & 0.6159 \\
 Vgg19 & 0.6414 & 0.6285 & 0.6405 \\
  \toprule
\end{tabular}
\caption{Segmentation Results using DeepLabV3}
\label{tab:DeepLab}
\end{center}
\end{table}

\begin{table}[h!]
\begin{center}
 \begin{tabular}{llll} 
 \toprule
 Encoder & $\text{mIoU}_\text{train}$ & $\text{mIoU}_\text{val}$ & $\text{mIoU}_\text{test}$ \\ [0.5ex]  
 \midrule
 RN18 & 0.7473 & 0.5803 & 0.5756 \\ 
 RN34 & 0.7342 & 0.5466 & 0.5404 \\
 RN50 & 0.7936 & 0.6813 & 0.6828 \\
 RN101 & 0.8086 & 0.4804 & 0.4627 \\
 RN152 & 0.7727 & 0.5058 & 0.5090 \\
 Vgg16 & 0.7180 & 0.7018 & 0.7131 \\
 Vgg19 & 0.7076 & 0.6864 & 0.6937 \\
 ENb0 & 0.7849 & 0.6828 & 0.6808 \\
 ENb1 & 0.7641 & 0.6208 & 0.6069 \\
 ENb2 & 0.7684 & 0.6382 & 0.6441 \\
 IncepV3 & 0.7607 & 0.5579 & 0.5514 \\
 IncepRNV2 & 0.7651 & 0.5927 & 0.5848 \\
 \toprule
\end{tabular}
\caption{Segmentation Results using PSPNet}
\label{tab:PSPNet}
\end{center}
\end{table}

For generating the baseline results and comparisons, mean intersection over union (mIoU) values are computed for training ($\text{mIoU}_{train}$), validation ($\text{mIoU}_{val}$) and testing set ($\text{mIoU}_{test}$). These mean IoU values also consist of the background class.

In Tables \ref{tab:Unet}, \ref{tab:LinkNet}, \ref{tab:PSPNet} and \ref{tab:DeepLab}, \textit{RN} means ResNet, \textit{ENb0} means EfficientNetb0 and similarly for \textit{ENb1}, \textit{ENb2}. \textit{IncepV3} means InceptionNetV3  and \textit{IncepRNV2} stands for InceptionResnetV2.

Table \ref{tab:Unet} shows the performance of Unet architecture with various encoders for semantic segmentation on the dataset (including the background class). Similarly, Tables \ref{tab:LinkNet}, \ref{tab:PSPNet} and \ref{tab:DeepLab} show the mean intersection over union values computed over the three splits (training, testing and validation) and for 12 classes (11 objects plus background).

\begin{figure}[h]
    \centering
    \includegraphics[width=0.50\textwidth]{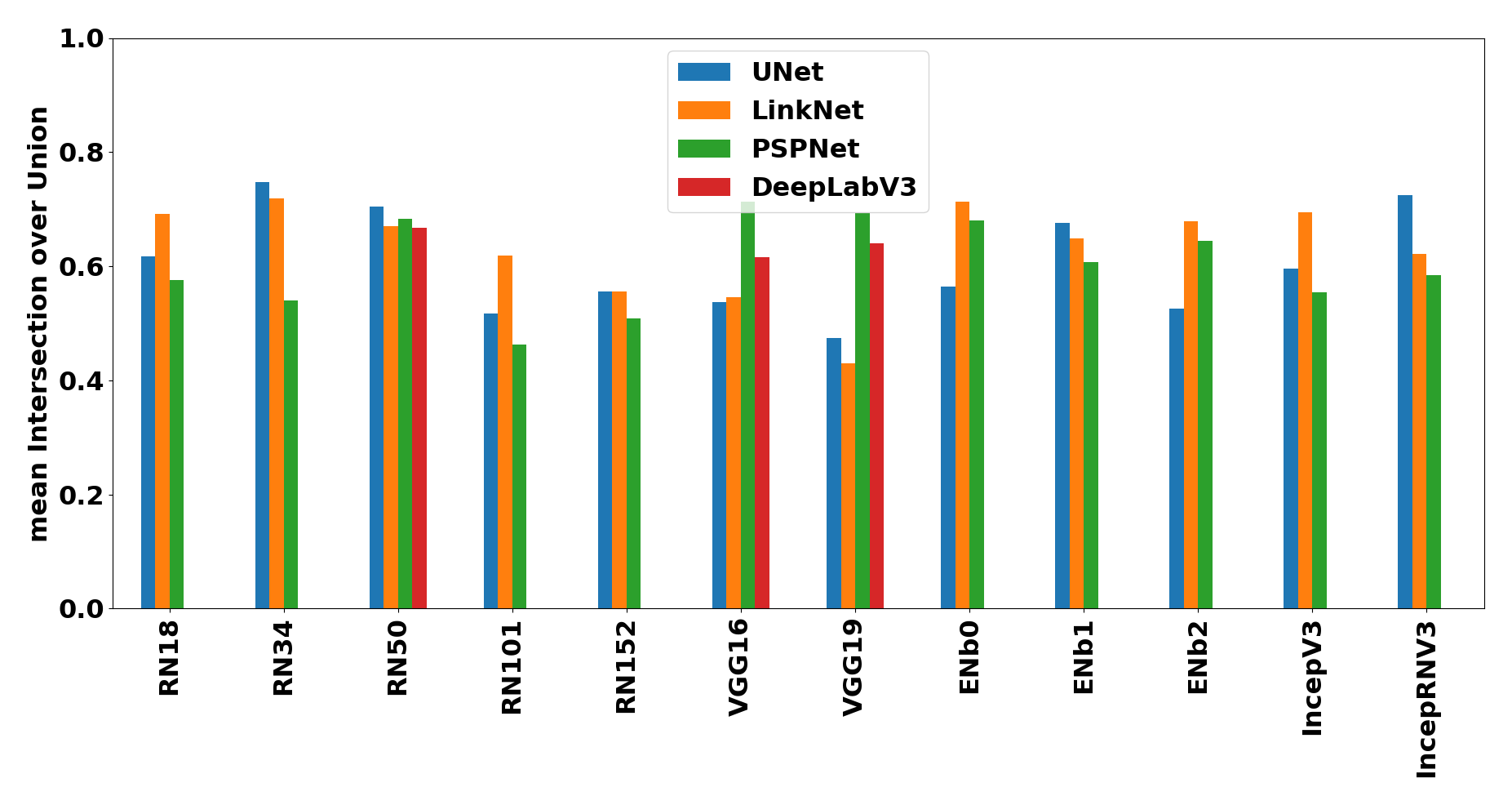}
    \caption{mIoU comparison for all classes (11 objects + background) on the test set}
    \label{fig:mIoU}
\end{figure}

Figure \ref{fig:mIoU} shows a comparison between the mean IoU values obtained on the test set for each segmentation model. Since for segmentation using deeplabv3, only resnet50, vgg16 and vgg19 encoders are used, there are only three bars for the same. 

\begin{figure*}[ht!]
\subfloat[Ground Truth]{
\includegraphics[width=0.19\textwidth]{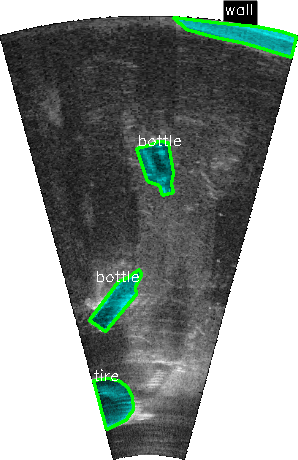}}
\subfloat[UNet+RN50]{
\includegraphics[width=0.19\textwidth]{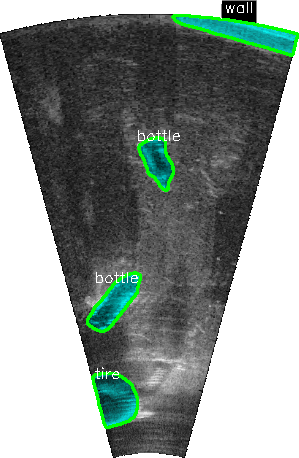}}
\subfloat[LinkNet+RN50]{
\includegraphics[width=0.19\textwidth]{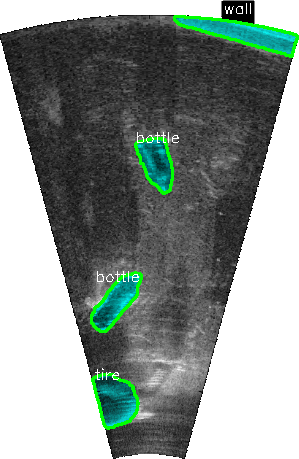}}
\subfloat[PSPNet+RN50]{
\includegraphics[width=0.19\textwidth]{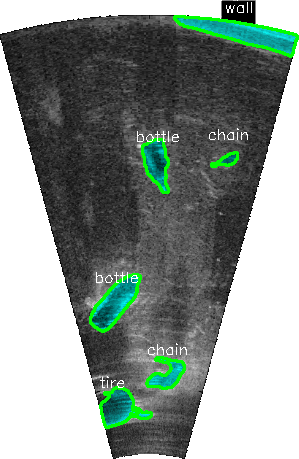}}
\subfloat[DeepLabV3+RN50]{
\includegraphics[width=0.19\textwidth]{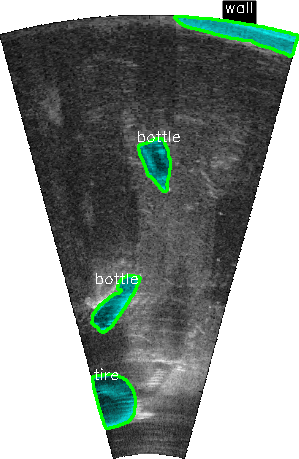}}
\caption{Segmentation outputs obtained using ResNet50 with Unet, LinkNet, Pspnet and DeepLabV3}
\label{fig:outR}
\end{figure*}

\begin{figure*}[ht!]
\subfloat[Ground Truth]{
\includegraphics[width=0.19\textwidth]{Images/Output/GT_MASK.png}}
\subfloat[UNet+VGG16]{
\includegraphics[width=0.19\textwidth]{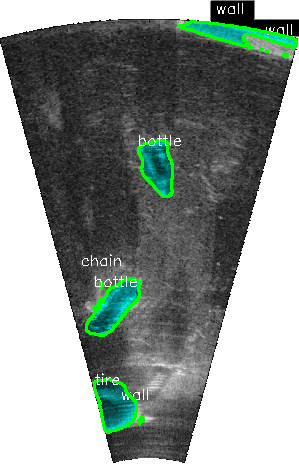}}
\subfloat[LinkNet+VGG16]{
\includegraphics[width=0.19\textwidth]{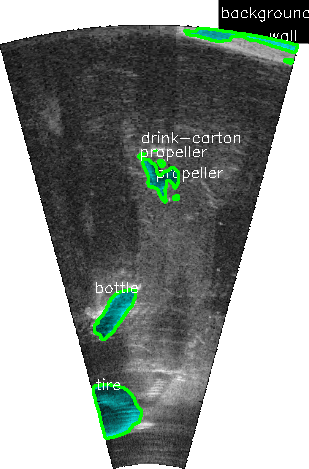}}
\subfloat[PSPNet+VGG16]{
\includegraphics[width=0.19\textwidth]{Images/Output/PSPResNet50_OverLay.png}}
\subfloat[DeepLabV3+VGG16]{
\includegraphics[width=0.19\textwidth]{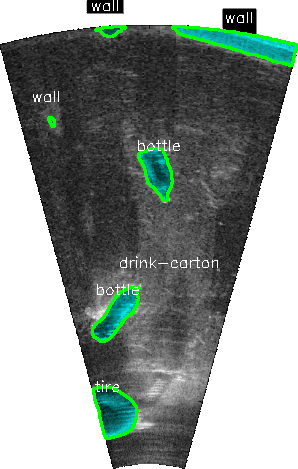}}
\caption{Segmentation outputs obtained using VGG16 with Unet, LinkNet, Pspnet and DeepLabV3}
\label{fig:outV}
\end{figure*}

Figures \ref{fig:outR} and \ref{fig:outV} present a comparison of segmentation results using the architectures described above with ResNet50 and VGG16 encoders. The class name corresponding to each object is mentioned on the image and the boundaries and areas enclosed within the boundaries are highlighted. As can be seen from these two figures, the encoder architecture has a significant effect on the output segmentation masks, but the segmentation architecture has a larger effect on varying mask quality.

The best performing model in our dataset according to test performance is Unet with ResNet34 backbone. Table \ref{tab:bestResults} presents per-class IoU on the test set by the top 4 best performing models.

\section{Conclusions and Future Work}

In this paper we presented a dataset for semantic segmentation of Forward-Looking Sonar images, in particular of marine debris objects in a synthetic water tank setting. We provide a large selection of baselines including different segmentation models and backbone architectures. We report that Unet with ResNet34 is the best performing model in the test set, with mean IoU of 0.7481.

Our dataset was captured in a limited setting, a water tank, and only limited views of the object are available (both in sensor pitch and object yaw). Data from a real underwater environment would be best, with different seafloor types (sand, rocky, etc), distractor objects (fish, algae), and marine growth.

We expect that our dataset and analysis motivates the community to use segmentation models in sonar images to increase the amount of information obtained from a computer vision, in comparison for classification and object detection tasks. The weights trained on our dataset can also be used as pre-trained weights for future tasks.

The dataset is publicly available at \url{https://github.com/mvaldenegro/marine-debris-fls-datasets/}.

\begin{table*}[!h]
    \begin{center}
        \begin{tabular}{lllll} 
            \toprule
            Object & $\text{Unet+RN34}$ & $\text{LinkNet+RN34}$ & $\text{PSPNet+VGG16}$ & $\text{DeepLabV3+RN50}$\\ [0.5ex]  
            \midrule
             background & \textbf{0.9926} & 0.9826 & 0.9857 & 0.9778 \\
            bottle & \textbf{0.7824} & 0.7350 & 0.7454 &  0.7121\\ 
            can & \textbf{0.6053} & 0.5709 & 0.5906 & 0.5396    \\
            chain & \textbf{0.6498} & 0.6046 & 0.6469 & 0.5880\\
            drink-carton & \textbf{0.6816} & 0.6072 & 0.6444 & 0.6215 \\ 
            hook & \textbf{0.6969} & 0.6592 & 0.6686 & 0.5335 \\ 
            propeller & 0.7089 & 0.6764 & \textbf{0.7185} & 0.6765 \\
            shampoo-bottle & 0.7288 & 0.7063 & \textbf{0.7361} & 0.5805 \\
            standing-bottle & 0.7998 & \textbf{0.8095} & 0.6721 & 0.6434 \\
            tire & \textbf{0.8869} & 0.8863 & 0.8736 & 0.8048 \\
            valve & \textbf{0.5613} & 0.5289 & 0.4711 & 0.5382 \\
            wall & 0.8829 & 0.8635 & \textbf{0.8771} & 0.7977 \\   
            \midrule
            mIoU & \textbf{0.7481} & 0.7192 & 0.7131 & 0.6678\\       
            \toprule
        \end{tabular}
        \caption{Per-Class IoU on the test set for best performing segmentation models.}
        \label{tab:bestResults}
    \end{center}
\end{table*}

{
\small
\bibliographystyle{ieee_fullname}
\bibliography{biblio}
}

\end{document}